\documentclass[runningheads]{llncs}

\usepackage[utf8]{inputenc}
\usepackage[T1]{fontenc}
\usepackage[english]{babel}
\usepackage{lmodern}

\usepackage{amsmath, amssymb}
\usepackage[ruled,onelanguage]{algorithm2e}
\usepackage{mathtools}

\usepackage{graphicx,wrapfig}
\usepackage{float}
\usepackage{subcaption}
\usepackage{natbib}
\bibpunct{(}{)}{;}{a}{,}{,}
\setcitestyle{numbers,open={[},close={]}}

\usepackage[svgnames]{xcolor}
\newcommand*{\doi}[1]{DOI: \href{http://dx.doi.org/\detokenize{#1}}{\detokenize{#1}}}
\usepackage{nameref}
\usepackage{hyperref}
\hypersetup{
  breaklinks,
  colorlinks,
  citecolor={blue},
  urlcolor={blue},
  linkcolor={blue}
}
\usepackage[nameinlink]{cleveref}

\usepackage[textwidth=\marginparwidth,
 textsize=scriptsize,
 colorinlistoftodos]{todonotes}
 
\newcommand{\dO}{{\partial\Omega}}  

\newcommand{\T}{^\mathrm{T}}       

\newcommand{\bv}{\ensuremath{\mathbf{b}}}

\newcommand{\gv}{\ensuremath{\mathbf{g}}}

\newcommand{\pv}{\ensuremath{\mathbf{p}}}

\newcommand{\sv}{\ensuremath{\mathbf{s}}}

\newcommand{\uv}{\ensuremath{\mathbf{u}}}

\newcommand{\zv}{\ensuremath{\mathbf{z}}}

\newcommand{\Fv}{\ensuremath{\mathbf{F}}}

\newcommand{\Nv}{\ensuremath{\mathbf{N}}}

\newcommand{\Wv}{\ensuremath{\mathbf{W}}}

\title{Real-time elastic partial shape matching using a neural network-based adjoint method}
\titlerunning{Real-time registration using a network-based adjoint method}
\author{Alban Odot \and Guillaume Mestdagh \and Yannick Privat \and Stéphane Cotin}
\authorrunning{A. Odot et al.}
\institute{Mimesis team, Inria, Strasbourg, France (\email{stephane.cotin@inria.fr})}
\date{November 2022}

\begin{document}

\maketitle

\begin{abstract}
Surface matching usually provides significant deformations that can lead to structural failure due to the lack of physical policy. In this context, partial surface matching of non-linear deformable bodies is crucial in engineering to govern structure deformations. In this article, we propose to formulate the registration problem as an optimal control problem using an artificial neural network where the unknown is the surface force distribution that applies to the object and the resulting deformation computed using a hyper-elastic model. The optimization problem is solved using an adjoint method where the hyper-elastic problem is solved using the feed-forward neural network and the adjoint problem is obtained through the backpropagation of the network. Our process improves the computation speed by multiple orders of magnitude while providing acceptable registration errors.

\end{abstract}

\section{Introduction}

We consider an elastic shape-matching problem between a deformable solid
and a point cloud.
Namely, an elastic solid in its reference configuration is represented by a
tridimensional mesh, while the point cloud represents a part of the solid boundary
in a deformed configuration.
The objective of the procedure is not only to deform the mesh so that
its boundary matches the point cloud, but also to estimate the displacement
field inside the object.

This situation also arises in computer-assisted liver surgery, where augmented reality
is used to help the medical staff navigate the operation scene \citep{haouchine2015}.
Most methods for intra-operative organ shape-matching revolve around a biomechanical model to
describe how the liver is deformed when forces are applied to its boundary.
Sometimes, a deformation is created by applying forces \citep{plantefeve2016}
or constraints \citep{peterlik2018,malti2015linear} to enforce surface correspondence.
Other approaches prefer to solve an inverse problem, where the final displacement
minimizes a cost functional among a range of admissible displacements
\citep{heiselman2020}.
However, while living tissues are known to exhibit a highly nonlinear behavior
\citep{marchesseau2017}, using hyperelastic models in the context of real-time
shape matching is prohibited due to high computational costs.
For this reason, the aforementioned methods either fall back to linear elasticity
\citep{heiselman2020} or to the linear co-rotational model \citep{plantefeve2016}.
In this paper, we perform real-time hyperelastic shape matching by predicting nonlinear
displacement fields using a neural network.
The network is included in an adjoint-like method, where the backward chain is
executed automatically using automatic differentiation.

Neural networks are used to predict solutions to partial differential equations,
in compressible aerodynamics~\citep{renganathan2021}, structural
optimization~\citep{white2019} or astrophysics~\citep{khan2021}.
Here we work at a small scale, but try to obtain real-time simulations using complex models.
Also, the medical image processing literature is full of networks that perform shape-matching
in one step \citep{pfeiffer2020}.
However, the range of available displacement fields is limited by the training dataset of the network, and thus less robust to unexpected deformations.
On the other hand, assigning a very generic task to the network results in a very flexible method, where details of the physical model, including the range of forces that can be applied to the liver and the zones where they apply may be chosen after the training.
Therefore, our shape-matching approach provides a good compromise between the speed of learning-based methods with the flexibility of standard simulations.
We want to mention that for the rest of this article due to how the method is formulated we interchangeably use the terms "shape-matching" and "registration".

We start by presenting the method split into three parts. First, the optimization problem; second, the used neural network and finally, the adjoint method computed using an automatic differentiation framework.

We then present the results considering a toy problem involving a square section beam and a more realistic one involving a liver.

\section{Methods}

\subsection{Optimization problem}
\begin{figure}[ht]
\centering
\includegraphics[width=0.6\linewidth]{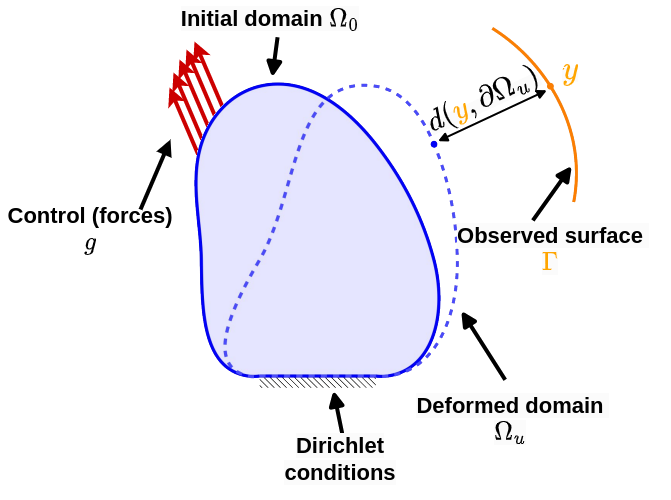}
\caption{Schematic of the problem which we are trying to optimize for.}\label{wrap-fig:toy_demo}
\vspace{-5ex}
\end{figure}

To model the registration problem, we use the optimal control formulation introduced  in \citet{mestdagh2022}.
The deformable object is represented by a tetrahedral mesh, endowed with a hyperelastic model.
In its reference configuration, the elastic object occupies the domain $\Omega_0$,
whose boundary is $\dO_0$.
When a displacement field $\uv$ is applied to $\Omega_0$, the deformed domain is
denoted by $\Omega_\uv$, and its boundary is denoted by $\partial \Omega_\uv$ as shown in \Cref{wrap-fig:toy_demo}.
Applying a surface force distribution $\gv$ onto the object boundary results in
the elastic displacement $\uv_\gv$, solution to the static equilibrium equation
\begin{equation}\label{eq:hyperelastic}
    \Fv(\uv_\gv) = \gv,
\end{equation}

where $\Fv$ is the residual from the hyperelastic model.
Displacements are discretized using continuous piecewise linear finite element functions
so that the system state is fully known through the displacement of mesh nodes, stored in~$\uv$.
Note that $\gv$ contains the nodal forces that apply on the mesh vertices.
As we only consider surface loadings, nodal forces are zero for nodes
inside the domain.
Finally, the observed data are represented by a point cloud $\Gamma = \{y_1, \dots, y_m\}$.
We compute a nodal force distribution that achieves the matching between $\dO_{\uv_\gv}$ and $\Gamma$ by solving the optimization problem
\begin{align}\label{eq:opt-problem}
    &\min_{\gv\in G}\quad \Phi(\gv) + \tfrac{\alpha}{2}\|\gv\|^2\\
    &\qquad\text{where}\qquad \Phi(\gv) = J(\uv_\gv),
\end{align}
where, $\alpha > 0$ is a regularization parameter, $G$ denotes the set of admissible nodal forces distributions, and $J$ is the least-square term
\begin{equation}\label{eq:functional}
    J(\uv) = \tfrac{1}{2m} \sum_{j=1}^m d^2(y_j, \dO_\uv).
\end{equation}

Here, $d(y, \dO_\uv) = \min_{x\in \partial \Omega_\uv} \|y - x\|$ denotes the distance
between $y\in\Gamma$ and $\dO_\uv$.
The functional $J$ measures the discrepancy between $\dO_\uv$ and $\Gamma$, and it
evaluates to zero whenever every point $y\in\Gamma$ is matched by $\partial\Omega_\uv$.

A wide range of displacement fields $\uv$ are minimizers of problem \eqref{eq:opt-problem},
but most of them have no physical meaning.
Defining a set of admissible controls $G$ is critical to generate only displacements that
are consistent with a certain physical scenario.
The set $B$ decides, among others, on which vertices nodal forces may apply,
but also which magnitude they are allowed to take.
Selecting zones where surface forces apply is useful to obtain
physically plausible solutions.

\subsection{A neural network to manage the elastic problem}
Nonlinear elasticity problems are generally solved using a Newton method, which yields very accurate displacement fields at a high computational cost. In this paper, we give a boost to the direct solution procedure by using a pre-trained neural network to compute displacements from forces. This results in much faster estimates, while the quality of solutions depends on the network training.

Artificial neural networks are composed of elements named artificial neurons grouped into multiple layers. A layer applies a transformation on its input data and passes it to the associated activation layer. The result of this operation is then passed to the next layer in the architecture.
Activation functions play an important role in the learning process of neural networks. Their role is to apply a nonlinear transformation to the output of the associated layers thus greatly improving the representation capacity of the network.

\begin{figure}[ht]
        \centering
        \includegraphics[width=.6\linewidth]{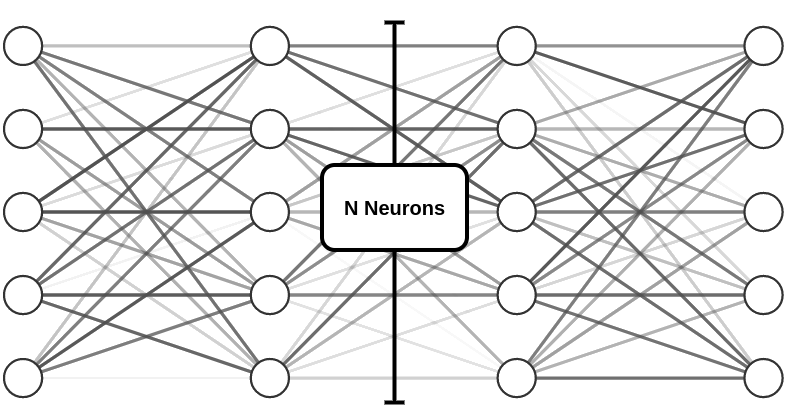}
        \caption{The proposed architecture is composed of 4 fully connected layers of size the number of degrees of freedom with a PReLU activation function. The input is the nodal forces and the output is the respective nodal displacements.}
        \label[figure]{fig:FCNN}
\end{figure}

While a wide variety of architectures are possible we will use the one proposed by \citet{odot2022}.
It consists of a fully-connected feed-forward neural network with 2 hidden layers
(see \Cref{fig:FCNN}).

The connection between two adjacent layers can be expressed as follows
\begin{equation}\label{eq:forward-propagation}
    \zv_i = \sigma_i(\Wv_i\zv_{i-1} + \bv_{i})\text{ for }1\leqslant i \leqslant n+1,
\end{equation}
where $n$ is the total number of layers, $\sigma(.)$ denotes the element wise activation function, $\zv_0$ and $\zv_{n+1}$ denotes the input and output tensors respectively, $\Wv_i$ and $\bv_i$ are the trainable weight matrices and biases in the $i^{th}$ layer.

In our case the activation functions $\sigma(.)$ are PReLU \cite{he2015}, which provides a learnable parameter $a$, allowing us to adaptively consider both positive and negative inputs.
From now on, we denote the forward pass operation in the network by
\begin{equation}
    \uv_\gv = \Nv(\gv).
\end{equation}

\subsection{An adjoint method involving the neural network}
We now give a closer look at the procedure to evaluate $\Phi$ and its derivatives.
We use an adjoint method, where the only variable controlled by the optimization solver is $\gv$.
As $J$ only operates on displacement fields,  the physical model plays the role of an intermediary
between these two protagonists.
The adjoint method is well suited to the network-based configuration, as the network can be used
as a black box.

In a standard adjoint procedure, a displacement is computed from a force distribution by solving
\eqref{eq:hyperelastic} using a Newton method, and it is then used to evaluate $\Phi(\gv)$. The Newton method is the algorithm of choice when dealing with non-linear materials, it iteratively solves the hyper-elastic problem producing accurate solutions. This method is also known for easily diverging when the load is reaching a certain limit that depends on the problem. To compute the deformation, one requires the application of multiple substeps of load which highly increases the computation times.
The backward chain requires solving an adjoint problem to evaluate the objective gradient, namely
\begin{equation}\label{eq:adjoint-system}
    \nabla \Phi(\gv) = \pv_\gv
    \quad\text{where}\quad
    \nabla \Fv(\uv_\gv)\T\pv_\gv = \nabla J(\uv_\gv).
\end{equation}
In \eqref{eq:adjoint-system}, the adjoint state $\pv_\gv$ is solution to a linear system involving the
hyperelasticity Jacobian matrix $\nabla \Fv(\uv_\gv)$.
When the network is used, however, the whole pipeline is much more straightforward, as the network
forward pass is only composed of direct operations.
The network-based forward and backward chains read
\begin{equation}
    \Phi(\gv) = J \circ \Nv(\gv)
    \quad\text{and}\quad
    \nabla \Phi(\gv) = \pv_\gv = \left[\nabla\Nv(\gv)\right]\T \nabla J(\uv_\gv),
\end{equation}
respectively.
On a precautionnary basis, let us take a brief look at the (linear) adjoint operator $\nabla \Nv(\gv)\T$.
When $\nabla \Nv(\gv)\T$ is applied, the information propagates backward in the network, following
the same wires as the forward pass.
The displacement gradient $\nabla J(\uv_\gv)$ is fed to the output tensor $\sv_{n+1}$ and the adjoint state is read at the network entry $\sv_0$.
In between, the relation between two layers is the adjoint operation to \eqref{eq:forward-propagation}.
It reads
\begin{equation}\label{eq:backward-propagation}
    \sv_{i-1} = \Wv_i\T\, \nabla\sigma_i(\Wv_i\zv_{i-1} + \bv_{i})\,\sv_i \quad \text{ for } \quad 1\leqslant i \leqslant n+1,
\end{equation}
where $\nabla\sigma_i(\Wv_i\zv_{i-1} + \bv_{i})$ is a diagonal matrix saved during the forward pass.

The network-based adjoint procedure is summarized in \Cref{alg:adjoint}, keeping in mind the backward chain is handled automatically.
Given a nodal force vector $\gv$, evaluating $\Phi(\gv)$ and $\nabla \Phi(\gv)$ requires one forward pass
and one backward pass in the network.
Then, \eqref{eq:opt-problem} may be solved iteratively using a standard gradient-based optimization
algorithm.
Because both network passes consist only of direct operations, the optimization solver is less likely to fail for accuracy reasons, compared to a $\Phi$ evaluation based on an iterative method.
\vspace{3ex}

\begin{algorithm}[H]
    \SetAlgoLined
    \KwData{Current iterate $\gv$}
    Perform the forward pass $\uv_\gv = \Nv(\gv)$\\
    Evaluate $J(\uv_gv)$ and $\nabla J(\uv_\gv)$\\
    Perform the backward pass $\pv_\gv = \left[\nabla\Nv(\gv)\right]\T \nabla J(\uv_\gv) $ \\
    \KwResult{$\nabla \Phi(\gv) = \pv_\gv$}
    \label[algorithm]{alg:adjoint}
    \caption{Network-based adjoint method to evaluate $\Phi$.}
\end{algorithm}


\section{Results}
\vspace{-1ex}
Our method is implemented in Python.
To be more specific, we use PyTorch to handle the network and evaluate $J$ on
the GPU, while the optimization solver is a limited memory BFGS algorithm 
\cite{byrd1995} available in the Scipy package.
Our numerical tests run on a Titan RTX GPU and AMD Ryzen 9 3950x CPU, with 32 GiB of RAM.

\subsection{Surface-matching tests on a beam mesh}\label[subsection]{subsec:31}
To assess the validity of our method, we first consider a toy problem involving a
square section beam with 304 hexahedal elements.
The network is trained using 20,000 pairs $(\gv,\uv_\gv)$, computed using
a Neo-Hookean material law with a Young modulus $E=4,500\ \mathrm{Pa}$
and a Poisson ratio $\nu = 0.49$.

\begin{figure}[H]
    \centering
    \begin{subfigure}[b]{0.32\linewidth}
        \centering
        \includegraphics[width=\linewidth]{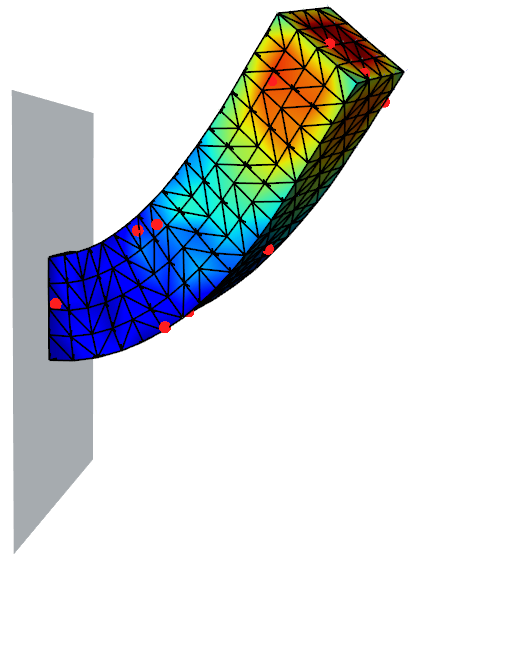}
        \caption{Reg. error: $5.9 \times 10^{-5}$,
        time: 0.07 s, iterations: 13}
        \label[figure]{fig:beam-bending}
    \end{subfigure}
\hfill
    \begin{subfigure}[b]{0.32\linewidth}
        \centering
        \includegraphics[width=\linewidth]{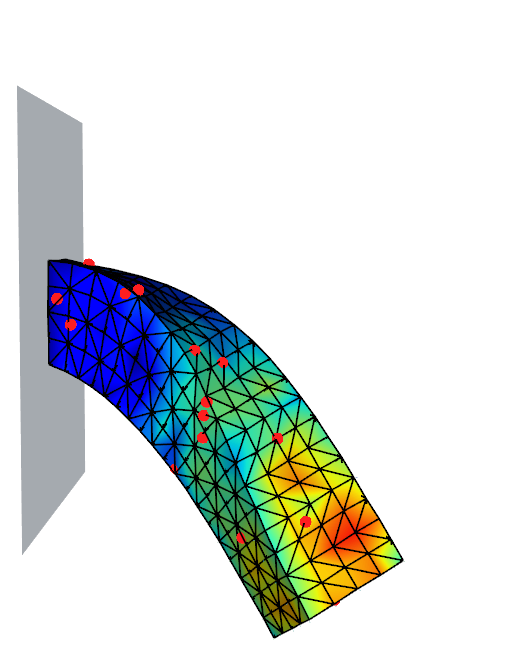}
        \caption{Reg. error: $6.6 \times 10^{-5}$ m, time: 0.09 s, iterations: 15}
        \label[figure]{fig:beam-both}
    \end{subfigure}
\hfill
    \begin{subfigure}[b]{0.32\linewidth}
        \centering
        \includegraphics[width=\linewidth]{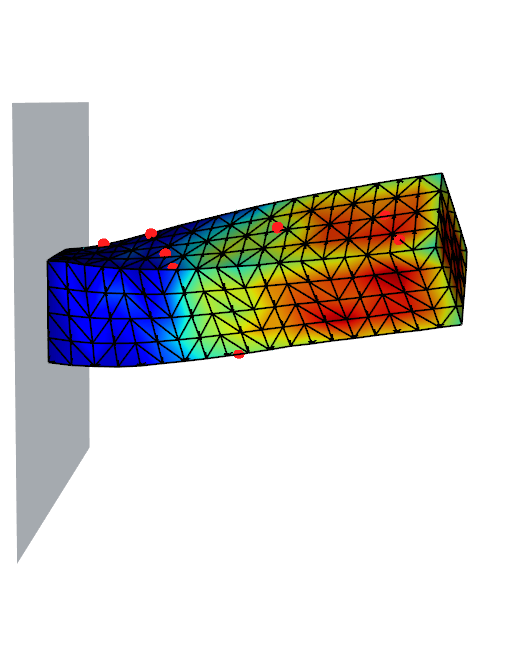}
        \caption{Reg. error: $3.4 \times 10^{-5}$,
        time: 0.115 s, iterations: 19}
        \label[figure]{fig:beam-torsion}
    \end{subfigure}
\caption{Deformations from the test dataset. 
         The red dots represent the target point clouds, and the color map represents the Von Mises stress error of the neural network prediction.}
\label[figure]{fig:beams_examples}
\end{figure}

We create 10,000 additional synthetic deformations of the beam, distinct from the
training dataset, using the SOFA finite element framework \cite{faure:hal-00681539}.
\Cref{fig:beams_examples} shows three examples of synthetic deformations, along with
the sampled point clouds.
Generated deformations include bending (\Cref{fig:beam-bending}), torsion (\Cref{fig:beam-torsion})
or a combination of them (\Cref{fig:beam-both}).
For each deformation, we sample the deformed surface to create a point cloud.
We then apply our algorithm with a relative tolerance of $10^{-4}$ on the objective gradient norm. We computed some statistics regarding the performance of our method over a series of 10,000 different scenarios and obtained the following results: mean registration error: $6 \times 10^{-5} \pm 6.15 \times 10^{-5}$, mean computation time: $48$ ms $\pm 19$ ms and mean number of iterations: 27 $\pm$ 11. 

Using a FEM solver, each sample of the test dataset took between 1 and 2 seconds to compute. This is mostly due to the complexity of the deformations as shown in \Cref{fig:beams_examples}. Such displacement fields require numerous costly Newton-Raphson iterations to reach equilibrium.
The neural network provides physical deformations in less than a millisecond regardless of the complexity of the force or resulting deformation, which highly improves the computation time of the method. From our analysis, the time repartition of the different tasks in the algorithm is pretty consistent, even with denser meshes. Network predictions and loss function evaluations represent $10\%$ to $15\%$ each, gradient computations represent up to the last $80\%$ of the whole optimization process.
This allows us to reach an average registration error of $5.37 \times 10^{-5}$ in less time than it takes to compute a single simulation of the problem using a classic FEM solver.

Due to the beam shape symmetry, some point clouds may be compatible with several
deformed configurations, resulting in wrong displacement fields returned by the procedure.
However, our procedure achieved a satisfying surface matching in each case.
These results on a toy scenario prove that our algorithm provides fast and accurate registrations.

In the next section, we apply our method in the field of augmented surgery with the partial surface registration of a liver and show that with no additional computation our approach produces with satisfying accuracy the forces that generate such displacements.

\subsection{An application in augmented surgery and robotics}
We now turn to another test case involving a more complex domain.
The setting is similar to \cite[Sect. 3.2]{mestdagh2022}.
In this context, a patient-specific liver mesh is generated from tomographic images and the objective is to provide augmented reality by registering, in real-time, the mesh to the deformed organ. During the surgery, only a partial point cloud of the visible liver surface can be obtained. The contact zones with the surgical instruments can also be estimated.
\begin{figure}
    \centering
    \includegraphics[width=0.75\linewidth]{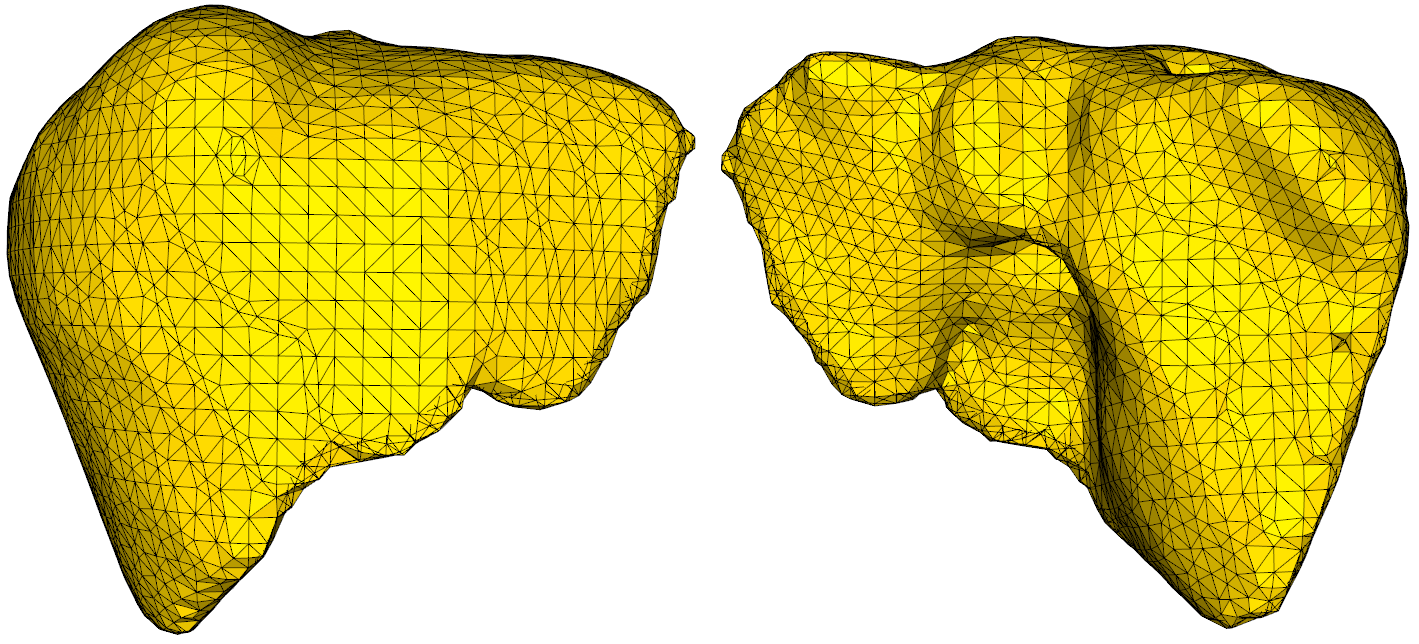}
    \caption{Mesh of the liver used in this section. Composed of 3,046 vertices and 10,703 tetrahedral elements which represents a challenge compared to the one used in \Cref{subsec:31}}
    \label{fig:Liver_at_rest}
\end{figure}
In our case, the liver mesh contains 3,046 vertices and 10,703 tetrahedral elements. Homogeneous Dirichlet conditions are applied at zones where ligaments hold the liver, and at the hepatic vein entry.
Like previously, we use a Neo-Hookean constitutive law with $E=4,500\ \mathrm{Pa}$ and $\nu = 0.49$, and the network is trained on 20,000 force/displacement pairs.
We create 5 series of synthetic deformations by applying a variable local force, distributed on a few nodes, on the liver mesh boundary.
For each series, 50 incremental displacements are generated, along with the corresponding point clouds.
The network-based registration algorithm is used to update the displacement field and forces between two frames.
We also run a standard adjoint method involving the Newton algorithm, to compare with our approach.
As the same mesh is used for data generation and reconstruction, the Newton-based reconstruction is expected to perform well.
\subsection{Liver partial surface matching for augmented surgery}

In this subsection, we present two relevant metrics: target registration error and computation times.
In augmented surgery, applications such as robot-aided surgery or holographic lenses require accurate calibrations that rely on registration. One of the most common metrics in registration tasks is the target registration error (TRE), which is the distance between corresponding markers not used in the registration process.
In our case we work on the synthetic deformation of a liver, thus, the markers will be the nodes of the deformed mesh. 
\begin{figure}[h]
    \centering
    \begin{subfigure}[b]{0.48\linewidth}
        \centering
        \includegraphics[width=\linewidth]{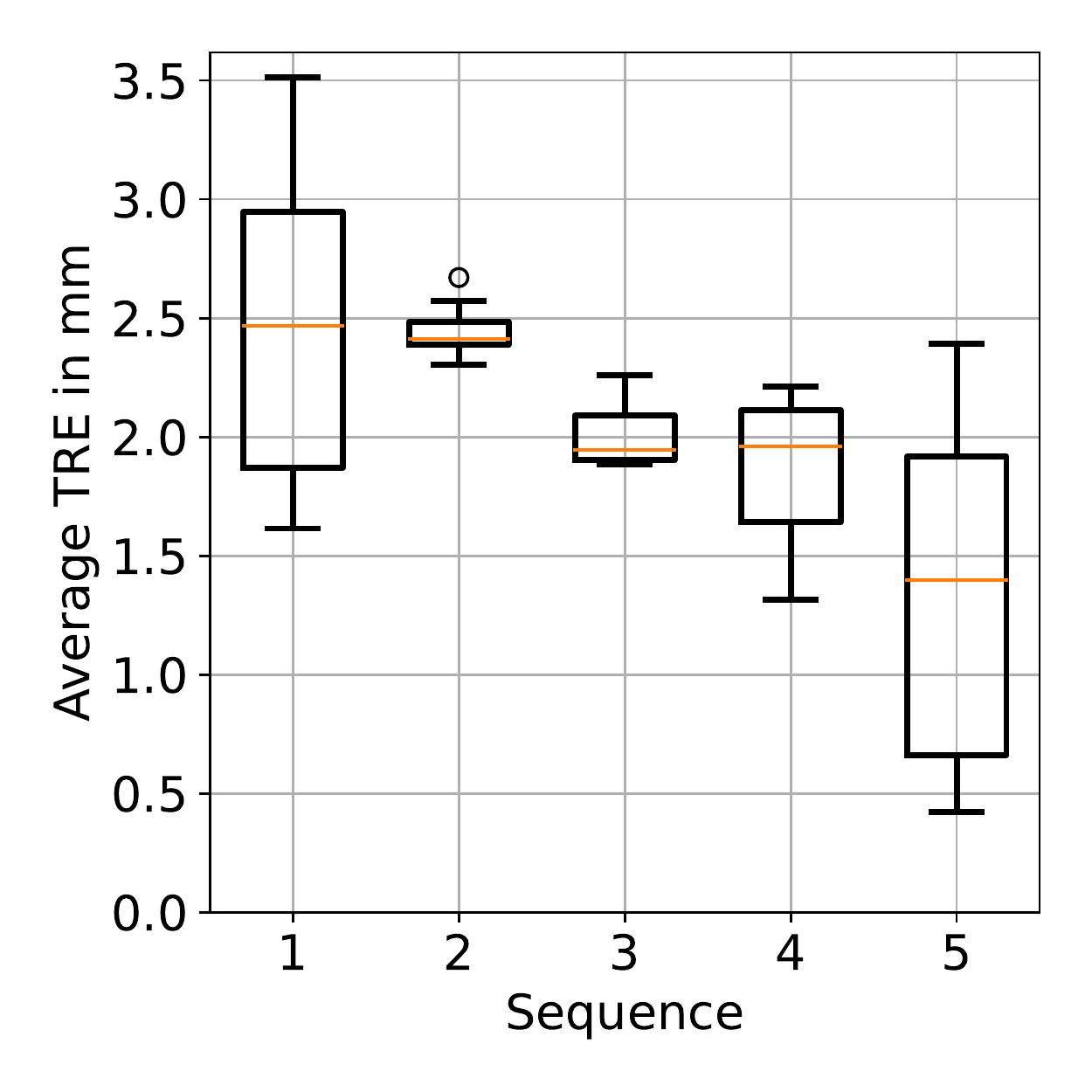}
         \label[figure]{fig:liver-tre}
    \end{subfigure}
    \begin{subfigure}[b]{0.48\linewidth}
        \centering
        \includegraphics[width=\linewidth]{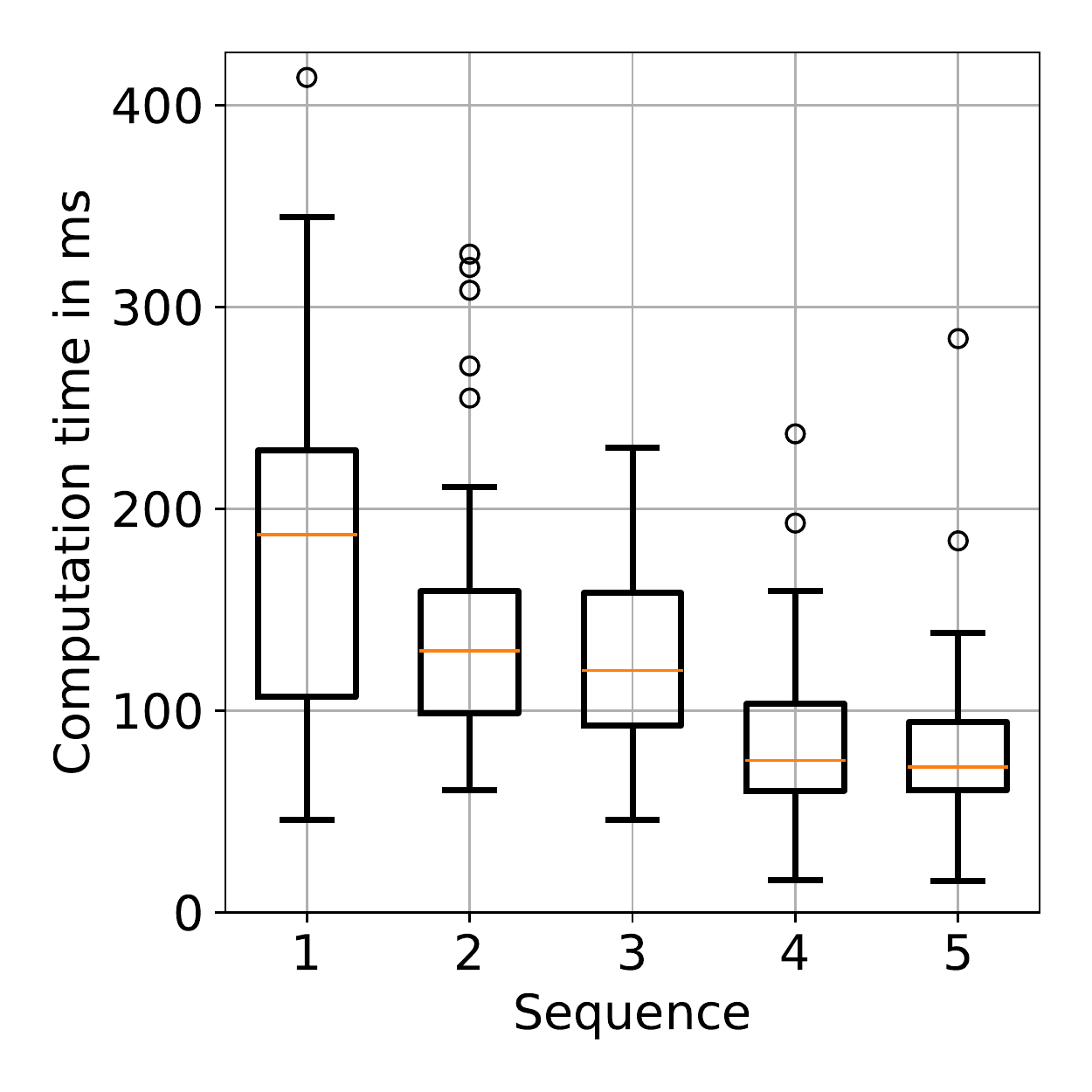}
         \label[figure]{fig:liver-times}
    \end{subfigure}
\caption{Average target registration error and computation times of each sequence.}
\label[figure]{fig:tre_times}
\end{figure}
The 5 scenarios present similar results with TRE between $3.5 \ mm$ and $0.5 \ mm$. Such errors are entirely acceptable and preserve the physical properties of the registered mesh. We point out that the average TRE for the classic method is around $0.1 \ mm$ which shows the impact of the network approximations.

Due to the non-linearity introduced by the Neo-Hookean material used to simulate the liver we need multiple iterations to converge toward the target point cloud. Considering the complexity of the mesh, computing a single iteration of the algorithm using a classical solver takes multiple seconds which leads to an average of 14 minutes per frame.
Our proposed algorithm uses a neural network to improve the computation speed of both the hyper-elastic and adjoint problems. The hyper-elastic problem takes around 4 to 5 milliseconds to compute while the adjoint problem takes around $11 \ ms$. This leads to great improvement in convergence speed as seen in \Cref{fig:liver-times} where on average we reduce the computation time by a factor of 6000.

\subsection{Force estimation for robotic surgery}
In the context of liver computer-assisted surgery, the objective is to estimate a force distribution supported by a small zone on the liver boundary.
Such a local force is for instance applied when a robotic instrument manipulates the organ.
In this case, it is critical to estimate the net force magnitude applied by the instrument, to avoid damaging the liver.
\begin{figure}[ht]
    \centering
    \begin{subfigure}[b]{\linewidth}
        \centering
        \includegraphics[width=\linewidth]{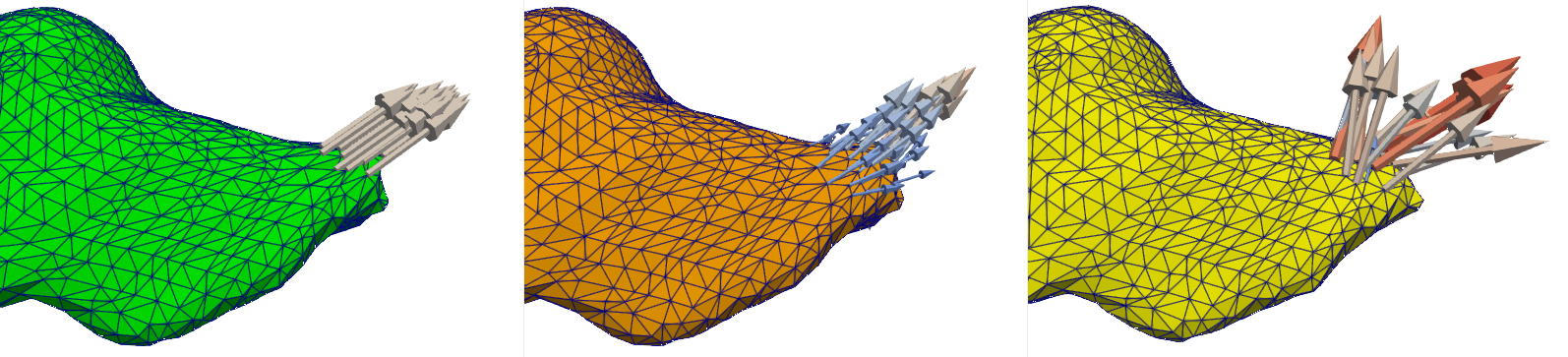}
        \caption{Frame 26}
    \end{subfigure}
    \begin{subfigure}[b]{\linewidth}
        \centering
        \includegraphics[width=\linewidth]{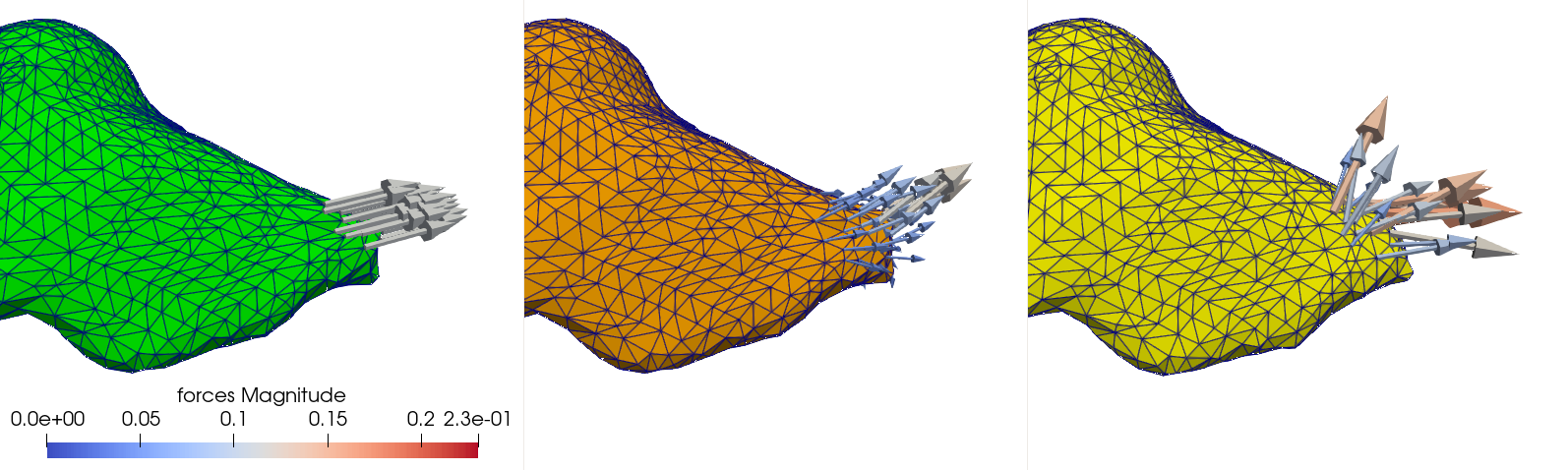}
        \caption{Frame 50}
    \end{subfigure}
\caption{Synthetic liver deformations and force distributions (left), reconstructed deformations
         and forces using the Newton method (middle) and the network (right) for test case 3.}
\label[figure]{fig:liver-compare}
\end{figure}
To represent the uncertainty about the position of the instruments the reconstructed forces are allowed to be nonzero on a larger support than the original distribution.
\Cref{fig:liver-compare} shows the reference and reconstructed deformations and nodal forces for three frames of the same series.
While the Newton-based reconstruction looks similar to the reference one, network-based
nodal forces are much noisier.
This is mostly due to the network providing only an approximation of the hyperelastic model.
\begin{wrapfigure}{r}{6.0cm}
    \centering
        \centering
        \includegraphics[width=\linewidth]{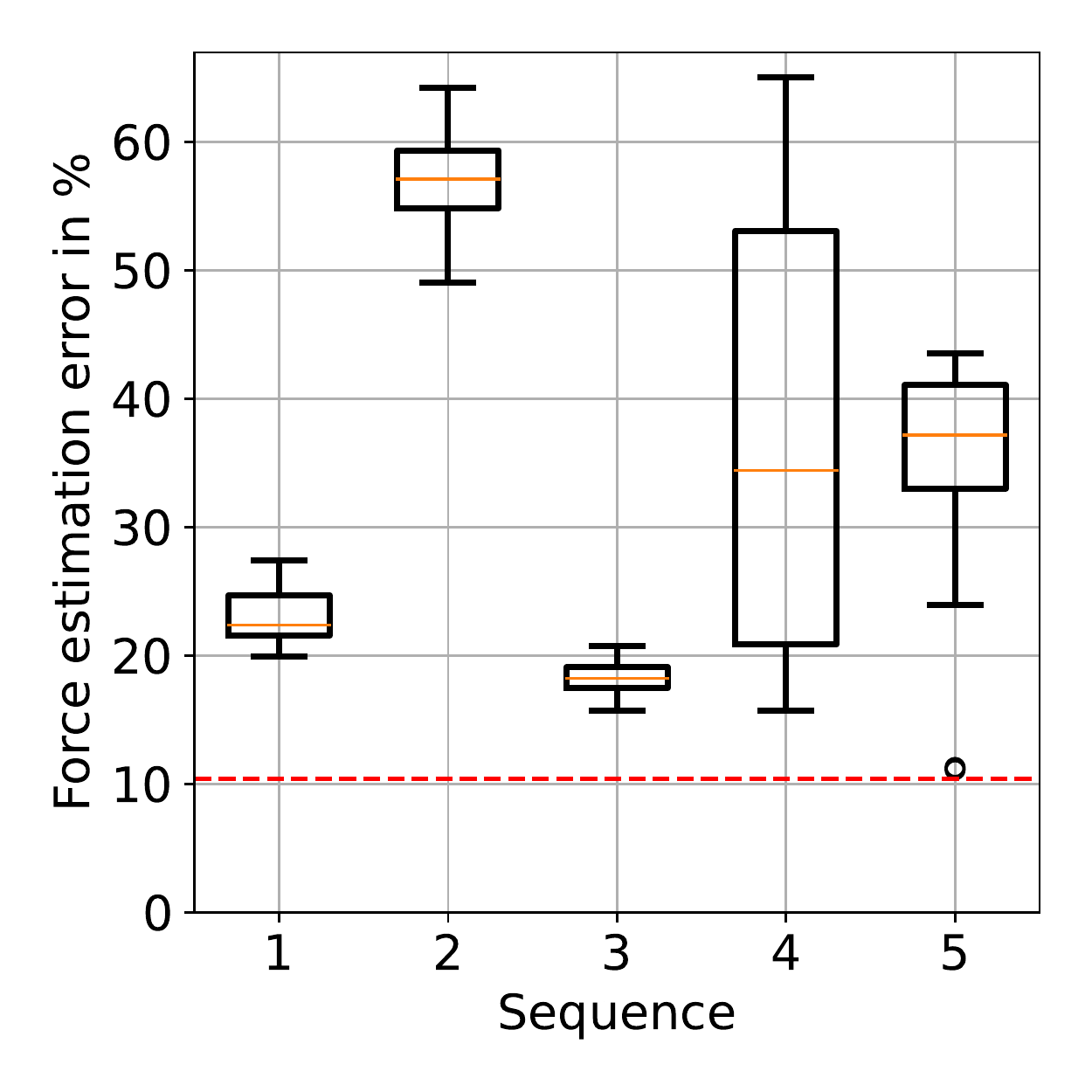}
        \caption{Force estimation error of the 5 sequences using our method, in red the average force reconstruction error with the classical method.}
    \label{fig:force_erros_seq}
    \vspace{-4ex}
\end{wrapfigure}
The great improvement in speed comes at the cost of precision. As shown in \Cref{fig:liver-compare} the neural network provides noisy force reconstructions. This is mostly due to prediction errors since the ANN only approximates solutions. These errors also propagate through the backward pass (adjoint problem), thus, accumulate in the final solution.
Although the force estimation is noisy for most cases it remains acceptable as displayed in \Cref{fig:force_erros_seq}. The red dotted line corresponds to the average error obtained with the classical adjoint method (10.04 \%). While we are not reaching such value, some sequences such as 1 and 3 provide good reconstructions. The difference in errors between scenarios is mostly due to training force distribution. This problem can be corrected by simply adding more data to the dataset thus providing better coverage of the force and deformation space.

These results show that this algorithm can produce fast and accurate registration at the expense of force reconstruction accuracy. This also shows that the force estimation is not directly correlated to registration accuracy. For example sequence 1 has the worst TRE but a good force reconstruction compared to sequence 4.

\section{Conclusion}
We presented a physics-based solution for a partial surface-matching problem that works with non-linear material using deep learning and optimal control formalism. The results are obtained on two main scenarios that differ both in scale and complexity. We showed that a fast and accurate registration can be obtained in both cases and can, in addition, predict the set of external forces that led to the deformation. Such results show that deep learning and optimal control have a lot in common and can be easily coupled to solve optimization problems very efficiently. Current limitations of our work are mostly due to the limited accuracy of the network and the need to retrain the network when the shape or material parameters of the model change.

\bibliographystyle{splncs04nat}
\bibliography{biblio}

\begin{thebibliography}{15}
\providecommand{\natexlab}[1]{#1}
\providecommand{\url}[1]{\texttt{#1}}
\providecommand{\urlprefix}{URL }
\expandafter\ifx\csname urlstyle\endcsname\relax
  \providecommand{\doi}[1]{doi:\discretionary{}{}{}#1}\else
  \providecommand{\doi}{doi:\discretionary{}{}{}\begingroup
  \urlstyle{rm}\Url}\fi

\bibitem[{Byrd et~al.(1995)Byrd, Lu, Nocedal, and Zhu}]{byrd1995}
Byrd, R.H., Lu, P., Nocedal, J., Zhu, C.: A limited memory algorithm for bound
  constrained optimization. SIAM Journal on Scientific Computing
  \textbf{16}(5), 1190--1208 (1995), \doi{10.1137/0916069}

\bibitem[{Faure et~al.(2012)Faure, Duriez, Delingette, Allard, Gilles,
  Marchesseau, Talbot, Courtecuisse, Bousquet, Peterlik, and
  Cotin}]{faure:hal-00681539}
Faure, F., Duriez, C., Delingette, H., Allard, J., Gilles, B., Marchesseau, S.,
  Talbot, H., Courtecuisse, H., Bousquet, G., Peterlik, I., Cotin, S.: {SOFA: A
  Multi-Model Framework for Interactive Physical Simulation}. In: {Soft Tissue
  Biomechanical Modeling for Computer Assisted Surgery}, Studies in
  Mechanobiology, Tissue Engineering and Biomaterials, vol.~11, pp. 283--321,
  Springer (2012), \doi{10.1007/8415\_2012\_125}

\bibitem[{{Haouchine} et~al.(2015){Haouchine}, {Cotin}, {Peterl{\'i}k},
  {Dequidt}, {Lopez}, {Kerrien}, and {Berger}}]{haouchine2015}
{Haouchine}, N., {Cotin}, S., {Peterl{\'i}k}, I., {Dequidt}, J., {Lopez}, M.S.,
  {Kerrien}, E., {Berger}, M.: Impact of soft tissue heterogeneity on augmented
  reality for liver surgery. IEEE Transactions on Visualization and Computer
  Graphics \textbf{21}(5), 584--597 (2015), \doi{10.1109/TVCG.2014.2377772}

\bibitem[{He et~al.(2015)He, Zhang, Ren, and Sun}]{he2015}
He, K., Zhang, X., Ren, S., Sun, J.: Delving deep into rectifiers: Surpassing
  human-level performance on imagenet classification. In: Proceedings of the
  IEEE International Conference on Computer Vision (ICCV) (2015)

\bibitem[{Heiselman et~al.(2020)Heiselman, Jarnagin, and Miga}]{heiselman2020}
Heiselman, J.S., Jarnagin, W.R., Miga, M.I.: Intraoperative correction of liver
  deformation using sparse surface and vascular features via linearized
  iterative boundary reconstruction. IEEE Transactions on Medical Imaging
  \textbf{39}(6), 2223--2234 (2020), \doi{10.1109/TMI.2020.2967322}

\bibitem[{Khan and Green(2021)}]{khan2021}
Khan, S., Green, R.: Gravitational-wave surrogate models powered by artificial
  neural networks. Phys. Rev. D \textbf{103}, 064015 (2021),
  \doi{10.1103/PhysRevD.103.064015}

\bibitem[{Malti et~al.(2015)Malti, Bartoli, and Hartley}]{malti2015linear}
Malti, A., Bartoli, A., Hartley, R.: A linear least-squares solution to elastic
  shape-from-template. In: Proceedings of the IEEE Conference on Computer
  Vision and Pattern Recognition, pp. 1629--1637 (2015)

\bibitem[{Marchesseau et~al.(2017)Marchesseau, Chatelin, and
  Delingette}]{marchesseau2017}
Marchesseau, S., Chatelin, S., Delingette, H.: Nonlinear biomechanical model of
  the liver. In: Payan, Y., Ohayon, J. (eds.) Biomechanics of Living Organs,
  Translational Epigenetics, vol.~1, pp. 243--265, Academic Press, Oxford
  (2017), \doi{10.1016/B978-0-12-804009-6.00011-0}

\bibitem[{Mestdagh and Cotin(2022)}]{mestdagh2022}
Mestdagh, G., Cotin, S.: An optimal control problem for elastic registration
  and force estimation in augmented surgery. In: Medical Image Computing and
  Computer Assisted Intervention -- MICCAI 2022, pp. 74--83, Springer Nature,
  Cham (2022), \doi{10.1007/978-3-031-16449-1_8}

\bibitem[{Odot et~al.(2022)Odot, Haferssas, and Cotin}]{odot2022}
Odot, A., Haferssas, R., Cotin, S.: {DeepPhysics}: A physics aware deep
  learning framework for real-time simulation. International Journal for
  Numerical Methods in Engineering \textbf{123}(10), 2381--2398 (2022),
  \doi{10.1002/nme.6943}

\bibitem[{Peterl{\'i}k et~al.(2018)Peterl{\'i}k, Courtecuisse, Rohling,
  Abolmaesumi, Nguan, Cotin, and Salcudean}]{peterlik2018}
Peterl{\'i}k, I., Courtecuisse, H., Rohling, R., Abolmaesumi, P., Nguan, C.,
  Cotin, S., Salcudean, S.: Fast elastic registration of soft tissues under
  large deformations. Medical Image Analysis \textbf{45}, 24--40 (2018), ISSN
  1361-8415, \doi{10.1016/j.media.2017.12.006}

\bibitem[{Pfeiffer et~al.(2020)Pfeiffer, Riediger, Leger, K{\"u}hn, Seppelt,
  Hoffmann, Weitz, and Speidel}]{pfeiffer2020}
Pfeiffer, M., Riediger, C., Leger, S., K{\"u}hn, J.P., Seppelt, D., Hoffmann,
  R.T., Weitz, J., Speidel, S.: Non-rigid volume to surface registration using
  a data-driven biomechanical model. In: Medical Image Computing and Computer
  Assisted Intervention -- MICCAI 2020, pp. 724--734, Springer International
  Publishing (2020)

\bibitem[{Plantef{\`e}ve et~al.(2016)Plantef{\`e}ve, Peterl{\'i}k, Haouchine,
  and Cotin}]{plantefeve2016}
Plantef{\`e}ve, R., Peterl{\'i}k, I., Haouchine, N., Cotin, S.:
  Patient-specific biomechanical modeling for guidance during
  minimally-invasive hepatic surgery. Annals of Biomedical Engineering
  \textbf{44}(1), 139--153 (2016), \doi{10.1007/s10439-015-1419-z}

\bibitem[{Renganathan et~al.(2021)Renganathan, Maulik, and
  Ahuja}]{renganathan2021}
Renganathan, S.A., Maulik, R., Ahuja, J.: Enhanced data efficiency using deep
  neural networks and gaussian processes for aerodynamic design optimization.
  Aerospace Science and Technology \textbf{111}, 106522 (2021),
  \doi{10.1016/j.ast.2021.106522}

\bibitem[{White et~al.(2019)White, Arrighi, Kudo, and Watts}]{white2019}
White, D.A., Arrighi, W.J., Kudo, J., Watts, S.E.: Multiscale topology
  optimization using neural network surrogate models. Computer Methods in
  Applied Mechanics and Engineering \textbf{346}, 1118--1135 (2019),
  \doi{10.1016/j.cma.2018.09.007}

\end{thebibliography}

\setcounter{tocdepth}{1}
\end{document}